\documentclass{article}

\usepackage{spconf,amsmath,graphicx,hyperref,booktabs,array,multirow,makecell,graphicx,arydshln}

\title{Cross-Modal Knowledge Distillation for Speech Large Language Models}

\name{Enzhi Wang\textsuperscript{1,2}, Qicheng Li\textsuperscript{1}*\thanks{* Corresponding author: liqicheng@nankai.edu.cn}, Zhiyuan Tang\textsuperscript{2}, Yuhang Jia\textsuperscript{1}}
\address{TMCC, College of Computer Science, Nankai University, Tianjin, China \\ Tencent Ethereal Audio Lab, Tencent Corporation, Shenzhen, China}

\begin{document}
\maketitle
\begin{abstract}
In this work, we present the first systematic evaluation of catastrophic forgetting and modality inequivalence in speech large language models, showing that introducing speech capabilities can degrade knowledge and reasoning even when inputs remain textual, and performance further decreases with spoken queries. To address these challenges, we propose a cross-modal knowledge distillation framework that leverages both text-to-text and speech-to-text channels to transfer knowledge from a text-based teacher model to a speech LLM. Extensive experiments on dialogue and audio understanding tasks validate the effectiveness of our approach in preserving textual knowledge, improving cross-modal alignment, and enhancing reasoning in speech-based interactions.
\end{abstract}

\section{Introduction}
\label{sec:intro}

In recent years, large language models (LLMs) have made remarkable progress in multimodal capabilities, with voice interaction emerging as a key application direction. Cutting-edge models such as GPT-4o \cite{Achiam2023} already enable real-time spoken dialogue, providing users with more natural, flexible, and high-quality interaction experiences compared to traditional text-based systems. Building on this trend, many researchers have begun extending pretrained text LLMs into the speech domain, constructing large speech models with both speech understanding and generation abilities. These models typically add a speech encoder and modality adapter layer on top of the text LLM \cite{Chu2024}\cite{Xu2025}\cite{Ding2025}\cite{wu2025step}, enabling the transformation from audio to text, thereby equipping the model with the ability to comprehend and respond to speech.  

However, the introduction of speech capabilities often leads to significant performance degradation. VoiceBench \cite{Chen2024} was among the first to observe that, under the same semantic input, the responses from speech input differ substantially from those of text input, suggesting that performance in pure speech interaction lags behind that in text mode. We approach this issue from the perspective of continual learning and argue that such models suffer from catastrophic forgetting \cite{McCloskey1989}. When adapting LLMs to speech tasks via continual learning, the model tends to overly focus on speech-related tasks while forgetting previously acquired language knowledge. This severely undermines the application potential of extending text-based LLMs into the speech modality.  

To mitigate catastrophic forgetting, many studies adopt a ``frozen backbone, adapter training'' strategy. For example, Freeze-Omni \cite{Wang2024} proposes to fully freeze the pretrained text LLM when introducing speech input, training only the speech encoder and adapter layers, thus preserving the ``intelligence'' level of the text model in speech mode. Similarly, Seed-ASR \cite{Bai2024} and FireRedASR-LLM \cite{Xu2025b} adopt an Encoder-Adapter-LLM architecture for end-to-end automatic speech recognition, also fixing the LLM while learning adapter layers to integrate speech and text capabilities. Although these approaches preserve the text model’s abilities to some extent, reasoning capabilities still tend to decline under the speech modality.  

Therefore, we attribute this performance degradation to two main factors: catastrophic forgetting and modality inequivalence. The latter often arises from poor alignment between the speech and text modalities, leading to performance drops. For instance, Freeze-Omni \cite{Wang2024} shows no forgetting under text input but exhibits large performance declines under speech input. Gong et al. highlight that in speech LLMs, acoustic and semantic representations often conflict: the acoustic objective (retaining high-quality audio features) and the semantic objective (accurately capturing textual meaning) are difficult to satisfy simultaneously under low bitrate or limited model capacity \cite{Gong2025}. In other words, the model faces a trade-off between preserving acoustic detail and comprehending audio content. EchoX  \cite{Zhang2025} further demonstrates that conventional training paradigms for speech LLMs fail to effectively bridge the gap between acoustic and semantic feature spaces, resulting in weaker reasoning and knowledge coverage in the speech modality compared to the text modality.

\begin{table*}[htbp]
\centering
\caption{Evaluation results on VoiceBench across text and speech modalities. T$\to$T: text-to-text performance, S$\to$T: speech-to-text performance. Note that we used AlpacaEval instead of AlpacaEval full, so some values differ from the official ones.}
\label{tab:voicebench}
\resizebox{0.95\textwidth}{!}{
\begin{tabular}{lcccccccc}
\toprule
\textbf{Model} & 
\multicolumn{2}{c}{\textbf{Open-Ended QA}} & 
\textbf{Knowledge} & 
\multicolumn{2}{c}{\textbf{Multi-Choice QA}} & 
\textbf{Instruction} & 
\textbf{Safety} & 
\textbf{Overall} \\
\cmidrule(lr){2-3} \cmidrule(lr){4-4} \cmidrule(lr){5-6} \cmidrule(lr){7-7} \cmidrule(lr){8-8}
 & AlpacaEval & CommonEval & SD-QA & MMSU & OpenBookQA & IFEval & AdvBench &  \\ 
\midrule
LLaMA-3.1-8B-Instruct & 4.69 & 4.38 & 76.59 & 66.23 & 72.53 & 76.72 & 96.54 & 81.43 \\
LLaMA-Omni (T$\to$T)  & 4.39 & 4.32 & 57.87 & 59.01 & 79.34 & 50.96 & 98.46 & 74.26 \\
LLaMA-Omni (S$\to$T)  & 3.70 & 3.46 & 39.69 & 25.93 & 27.47 & 14.87 & 11.35 & 37.51 \\
\midrule
Qwen2-0.5B-Instruct   & 2.95 & 2.95 & 22.25 & 30.20 & 36.04 & 21.83 & 95.39 & 46.24 \\
Mini-Omni2 (T$\to$T)  & 2.65 & 2.86 & 11.39 & 27.13 & 32.09 & 14.01 & 92.88 & 41.10 \\
Mini-Omni2 (S$\to$T)  & 2.32 & 2.18 &  9.31 & 24.27 & 26.59 & 11.56 & 57.50 & 31.32 \\
\midrule
Qwen-7B-chat          & 4.45 & 4.01 & 58.24 & 45.56 & 69.45 & 35.85 & 99.80 & 68.30 \\
Qwen2-Audio (T$\to$T) & 4.11 & 3.77 & 51.17 & 45.02 & 67.91 & 33.38 & 96.73 & 64.54 \\
Qwen2-Audio (S$\to$T) & 3.74 & 3.43 & 35.71 & 35.72 & 49.45 & 26.33 & 96.73 & 55.35 \\
\midrule
Qwen2-7B-Instruct     & 4.87 & 4.43 & 64.92 & 66.55 & 82.86 & 56.43 & 98.27 & 79.14 \\
Freeze-Omni (T$\to$T) & -    & -    & -     & -     & -     & -     & -     & -    \\
Freeze-Omni (S$\to$T) & 4.03 & 3.46 & 53.45 & 28.14 & 30.98 & 23.40 & 97.30 & 54.72 \\
\midrule
Qwen2.5-7B-Instruct   & 4.84 & 4.45 & 68.45 & 71.65 & 83.74 & 71.15 & 98.85 & 82.81 \\
Minicpm (T$\to$T)     & 4.47 & 4.10 & 63.11 & 59.42 & 82.86 & 47.73 & 98.84 & 74.77 \\
Minicpm (S$\to$T)     & 4.39 & 4.15 & 50.72 & 54.78 & 78.02 & 45.25 & 97.69 & 71.04 \\
Qwen2.5-Omni (T$\to$T)& 4.61 & 4.24 & 61.39 & 67.94 & 84.40 & 59.70 & 99.80 & 78.60 \\
Qwen2.5-Omni (S$\to$T)& 4.60 & 3.98 & 58.23 & 61.51 & 81.09 & 53.33 & 99.80 & 75.08 \\
\bottomrule
\end{tabular}}
\end{table*}

To tackle these challenges, we conduct a systematic study with three main contributions:  

\textbf{(1) General validation.} We perform the first systematic evaluation of multiple mainstream open-source speech LLMs regarding catastrophic forgetting and modality inequivalence. By feeding the same text queries to both a speech LLM and its corresponding text base model, we quantify the degradation of reasoning ability in text mode. Meanwhile, by comparing the outputs of speech queries and text queries on the same speech LLM, we assess inequivalence across modalities. Experiments confirm that these problems are widespread.  

\textbf{(2) Cross-modal knowledge distillation.} We propose a novel strategy to jointly address forgetting and inequivalence. Inspired by cross-modal distillation in computer vision \cite{Huo2024}\cite{Ji2024}, we treat the speech LLM as the student and its text LLM backbone (with stronger reasoning and knowledge capabilities) as the teacher. Distillation is performed in two stages: (a) \textit{text-to-text distillation}: using open instruction datasets such as OpenOrca \cite{Mukherjee2023}, the speech LLM is trained to mimic the teacher’s outputs under text input, alleviating catastrophic forgetting; (b) \textit{speech-to-text distillation}: the same text data are converted to synthetic speech via TTS and fed to the speech LLM, while the teacher outputs remain the supervision signal, thereby strengthening semantic alignment across modalities. Prior works such as Desta2.5 \cite{Lu2025} and MiDashengLM \cite{Dinkel2025} explored leveraging text LLM knowledge via instruction tuning to enhance speech LLMs in acoustic audio analysis tasks. Their method demonstrated that text-based knowledge can improve audio modality performance in tasks such as ``listing possible sound sources,'' and ``generate a descriptive caption for this audio.''. Or like OSUM-EChat \cite{Geng2025}, input the emotion label into the text LLM to obtain empathy information. 

However, such work remains largely limited to acoustic information and audio analysis. Unlike vision, audio in speech LLMs is not only an object of analysis but also serves as the query modality in dialogue tasks. For example, VoxEval \cite{Cui2025} evaluates mathematical problem-solving via spoken questions, and VoiceBench \cite{Chen2024} benchmarks speech-to-text dialogue tasks. To this end, we are the first to apply explicit cross-modal semantic knowledge distillation to direct dialogue tasks (text-to-text and speech-to-text), which targets higher-level knowledge transfer and reasoning enhancement.

\textbf{(3) Experimental validation.} Using only about 60,000 samples, our method substantially improves the overall capabilities of speech LLMs such as Qwen2.5-Omni \cite{Xu2025}. Experimental results show that, after distillation, the model demonstrates clear gains in general ability tests, commonsense reasoning, multidisciplinary knowledge, instruction following, and complex audio analysis and reasoning tasks.

\section{Capability Degradation Evaluation}
\begin{figure*}[htbp] 
    \centering
    \includegraphics[width=0.75\textwidth]{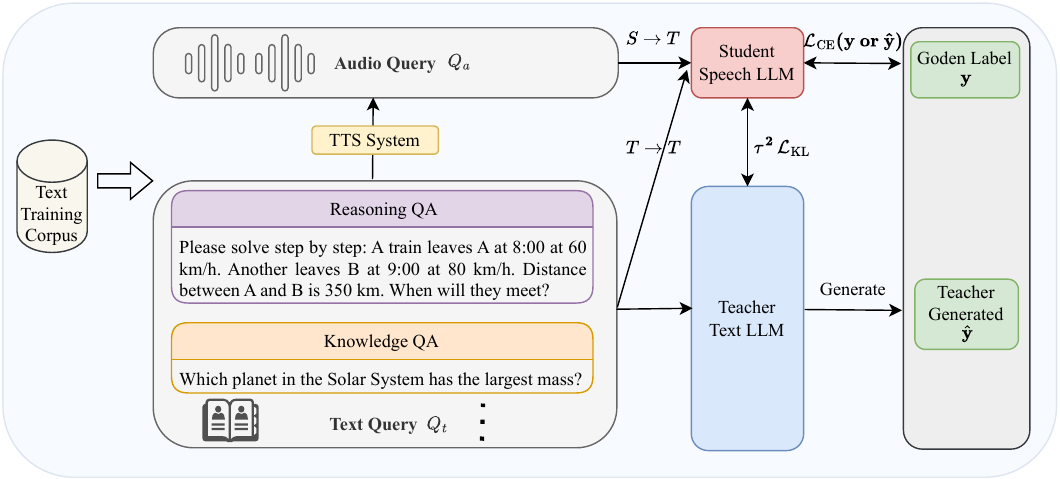} 
    \caption{An overview of the proposed \textbf{Cross-Modal Knowledge Distillation for Speech LLMs}, which consists of two complementary channels: \textbf{Speech-to-Text (S$\to$T)} and \textbf{Text-to-Text (T$\to$T)}. Each channel trains the speech LLM using the $\mathcal{L}_{\mathrm{CE}}$ with either the gold label $\mathbf{y}$ or the text LLM-generated label $\hat{\mathbf{y}}$, combined with the $\mathcal{L}_{\mathrm{KL}}$ scaled by the temperature $\tau^2$.} 
    \label{fig:framework} 
\end{figure*}
\subsection{Problem Definition}
We define the evaluation setting as follows: let the text-form question in a dialogue QA task be $Q_t$, and the corresponding speech-form question be $Q_a$. The text base model is denoted as $\theta_t$, and the speech large language model as $\theta_s$. Their corresponding outputs are:  
\[
T1 = \theta_t(Q_t), \quad
T2 = \theta_s(Q_t), \quad
T3 = \theta_s(Q_a).
\]  
The gap between $T1$ and $T2$ is defined as the \textbf{catastrophic forgetting}, while the performance gap between $T2$ and $T3$ is defined as the \textbf{modality inequivalence}.  

\subsection{Evaluation Results}
We evaluate on the VoiceBench dataset, which provides both speech and text modalities across multiple QA dimensions. The results are shown in Table \ref{tab:voicebench}. Speech LLMs generally suffer from catastrophic forgetting, as their performance under text input is consistently lower than that of their corresponding text base models. For instance, LLaMA-Omni achieves 74.26 under text input, compared with 81.43 for LLaMA-3.1-8B-Instruct. 

These models also exhibit modality inequivalence, with speech input producing weaker results than text input. Although Freeze-Omni avoids catastrophic forgetting by freezing the backbone LLM, it performs poorly under speech input, highlighting a trade-off between preserving text knowledge and achieving cross-modal consistency. Overall, these results reveal that catastrophic forgetting and modality inequivalence are prevalent challenges in current speech LLMs,

\section{Method}
\label{sec:method}

\subsection{Notation and overview}
We denote the text-form question by $Q_t$ and the synthesized speech question by $Q_a$. The text-base (teacher) model is $\theta_t$ and the speech large language (student) model is $\theta_s$. For an input $x$ (text or speech) the teacher and student produce token-level logits sequences
\[
\mathbf{z}^T = \theta_t(Q_t),\qquad
\mathbf{z}^S = \theta_s(x),
\]
where $\mathbf{z}^T = \{z^T_1,\dots,z^T_{T}\}$ and $\mathbf{z}^S = \{z^S_1,\dots,z^S_{T'}\}$ are the logits at each decoding step (lengths $T$ and $T'$ may differ). We write the reference (gold) response as $\mathbf{y} = (y_1,\dots,y_{L})$ and the teacher-generated response as $\mathbf{\hat y} = (\hat y_1,\dots,\hat y_{\hat L})$. A TTS system $\mathcal{T}(\cdot)$ is used to synthesize speech from text: 
\[
Q_a = \mathcal{T}(Q_t).
\]

Our objective is to distill the teacher's knowledge into the student through two complementary channels:  
(i) \textbf{Text-to-Text (T$\to$T) distillation}, where the student is trained on textual inputs ($x=Q_t$) to mitigate catastrophic forgetting;  
(ii) \textbf{Speech-to-Text (S$\to$T) distillation}, where the student is trained on synthesized speech ($x=Q_a$) to align speech-mode outputs with the teacher’s text-mode outputs, reducing modality inequivalence.  

\begin{table*}[htbp]
\centering
\caption{Speech-to-Text (S$\to$T) evaluation results on VoiceBench. CE denotes cross-entropy training on dataset labels $\mathbf{y}$, while Teacher CE uses text LLM-generated labels $\hat{\mathbf{y}}$ as hard targets.}
\label{tab:stot_results}
\resizebox{0.90\textwidth}{!}{
\begin{tabular}{llcccccccc}
\toprule
\textbf{Method} & \textbf{Model} & 
\multicolumn{2}{c}{\textbf{Open-Ended QA}} & 
\textbf{Knowledge} & 
\multicolumn{2}{c}{\textbf{Multi-Choice QA}} & 
\textbf{Instruction} & 
\textbf{Safety} & 
\textbf{Overall} \\
\cmidrule(lr){3-4} \cmidrule(lr){5-5} \cmidrule(lr){6-7} \cmidrule(lr){8-8} \cmidrule(lr){9-9}
 &  & AlpacaEval & CommonEval & SD-QA & MMSU & OpenBookQA & IFEval & AdvBench &  \\ 
\midrule
\multirow{1}{*}{Base} 
    & Qwen2.5-Omni               & 4.60 & 3.98 & 58.23 & 61.51 & 81.09 & 53.33 & \textbf{99.80} & 75.08 \\
\hline
\multirow{4}{*}{S2T KD} 
    & CE                         & 4.20 & 3.78 & 53.44 & 58.81 & 80.88 & 47.39 & 98.65 & 71.25 \\
    & CE + KL                    & 4.45 & 3.98 & 56.87 & 62.71 & 79.34 & 53.47 & 99.80 & 74.40 \\
    & Teacher CE                 & 4.64 & 4.09 & 59.32 & 61.82 & 81.09 & 53.56 & 99.23 & 75.66 \\
    & Teacher CE + KL            & 4.57 & 4.10 & 59.50 & 62.84 & 79.56 & 55.61 & 99.42 & 75.76 \\
\hline
\multirow{1}{*}{S2T KD + T2T KD} 
    & Teacher CE                 & \textbf{4.66} & \textbf{4.14} & \textbf{60.94} & \textbf{63.09} & \textbf{82.64} & \textbf{58.40} & 99.23 & \textbf{77.19} \\
\bottomrule
\end{tabular}}
\end{table*}

\begin{table*}[htbp]
\centering
\caption{Text-to-Text (T$\to$T) evaluation results on VoiceBench. All results are obtained using the Teacher CE approach.}
\label{tab:tott_results}
\resizebox{0.87\textwidth}{!}{
\begin{tabular}{lcccccccc}
\toprule
\textbf{Method} & 
\multicolumn{2}{c}{\textbf{Open-Ended QA}} & 
\textbf{Knowledge} & 
\multicolumn{2}{c}{\textbf{Multi-Choice QA}} & 
\textbf{Instruction} & 
\textbf{Safety} & 
\textbf{Overall} \\
\cmidrule(lr){2-3} \cmidrule(lr){4-4} \cmidrule(lr){5-6} \cmidrule(lr){7-7} \cmidrule(lr){8-8}
 & AlpacaEval & CommonEval & SD-QA & MMSU & OpenBookQA & IFEval & AdvBench &  \\ 
\midrule
Base             & 4.61 & 4.24 & 61.39 & 67.94 & 84.40 & 59.70 & \textbf{99.80} & 78.60 \\
S2T KD           & 4.75 & 4.30 & 62.39 & 68.37 & 83.74 & 57.50 & 99.42 & 78.95 \\
S2T KD + T2T KD  & \textbf{4.75} & \textbf{4.31} & \textbf{63.20} & \textbf{69.15} & \textbf{84.62} & \textbf{61.60} & 99.42 & \textbf{79.86} \\
\bottomrule
\end{tabular}}
\end{table*}

\begin{table}[htbp]
\centering
\caption{Audio analysis reasoning performance on MMAU-mini (Original). Evaluation was performed using the Kimi-audio toolkit \cite{ding2025kimi} with GPT for automatic judgment instead of regex extraction, which differs from the official one.}
\label{tab:audio_results}
\resizebox{0.42\textwidth}{!}{
\begin{tabular}{lcccc}
\toprule
\textbf{Method} & \textbf{Music} & \textbf{Sound} & \textbf{Speech} & \textbf{Avg.} \\
\midrule
Base             & \textbf{70.36} & 81.38 & 70.87 & 74.20 \\
\hline
S2T KD           & 68.86 & 81.08 & \textbf{74.77} & 74.90 \\
\hline
S2T KD + T2T KD  & 69.16 & 80.48 & 73.27 & 74.30 \\
S2T KD + T2T KD + AQA & 68.01 & \textbf{84.08} & \textbf{74.77} & \textbf{78.95} \\
\bottomrule
\end{tabular}}
\end{table}
\subsection{Distillation losses}

\subsubsection{Text-to-Text Knowledge Distillation (T2T KD)}
In this setting, the student takes textual input $Q_t$ and is trained to reproduce either the gold labels $\mathbf{y}$ or teacher outputs $\hat{\mathbf{y}}$. The loss combines cross-entropy with the chosen hard targets and a Kullback Leibler divergence (KL) loss between teacher and student softened distributions with a temperature $\tau>0$:

\begin{equation*}
\mathcal{L}_{\mathrm{T\to T}} \;=\; \mathcal{L}_{\mathrm{CE}}(\mathbf{y}\ \text{or}\ \hat{\mathbf{y}};\, Q_t, \theta_s) \;+\; \lambda \tau^2 \mathcal{L}_{\mathrm{KL}}(Q_t;\, \theta_t,\theta_s).
\end{equation*}

\subsubsection{Speech-to-Text Knowledge Distillation (S2T KD)}
The student takes synthesized speech $Q_a=\mathcal{T}(Q_t)$ while the teacher uses the original text $Q_t$. The loss therefore, pairs the student conditioned on $Q_a$ with the teacher conditioned on $Q_t$:
\begin{equation*}
\mathcal{L}_{\mathrm{S\to T}}
\;=\; \mathcal{L}_{\mathrm{CE}}(\mathbf{y}\ \text{or}\ \hat{\mathbf{y}};\, Q_a, \theta_s)
\;+\; \lambda\,\tau^2\,\mathcal{L}_{\mathrm{KL}}(Q_t, Q_a;\, \theta_t,\theta_s).
\end{equation*}

\section{EXPERIMENTS AND RESULTS}
\label{sec:evaluation}

\subsection{Experiment Setup and Details}
We use Qwen2.5-Omni \cite{Xu2025} as the backbone LLM and Qwen2.5-7B-Instruct \cite{qwen2025qwen25technicalreport} as the text teacher LLM. Speech questions are synthesized using CosyVoice 2 \cite{du2024cosyvoice}. For distillation, we leverage the Open-Orca \cite{Mukherjee2023} dataset, consisting of 22,456 T$\to$T samples and 44,753 S$\to$T samples. All models are trained for 2 epochs with a learning rate of $5\times 10^{-6}$, 
the KL weight $\lambda$ is set to 0.5 and the temperature $\tau$ is set to 2.

\subsection{Speech-to-Text (S$\to$T) performance}
Table~\ref{tab:stot_results} presents the S$\to$T evaluation results. Training the student with simple cross-entropy on the dataset labels (CE) can even degrade performance, likely because the model requires a large and diverse dataset to learn effectively. Adding the KL divergence term (CE + KL) clearly improves CE, helping the student better match the teacher's distribution. However, using teacher-generated outputs as hard targets (Teacher CE) is more effective, providing high-quality guidance for reasoning and knowledge transfer. Incorporating both Teacher CE and KL (Teacher CE + KL) offers marginal additional gains. Finally, combining S2T KD with T2T KD (Teacher CE, S2T + T2T) achieves the best overall results, showing that preserving text-mode knowledge through T2T distillation complements speech-mode alignment and effectively mitigates catastrophic forgetting while enhancing speech understanding.

\subsection{Text-to-Text (T$\to$T) performance}
As shown in Table~\ref{tab:tott_results}, the T$\to$T distillation benefits from S2T knowledge distillation as well. Using teacher outputs from S2T KD slightly improves Open-Ended QA and Multi-Choice QA scores. When combined with T2T KD, the model achieves the highest overall performance, indicating that even in text-mode, incorporating cross-modal knowledge distillation can strengthen reasoning and knowledge capabilities.

\subsection{Audio analysis reasoning performance}
Table~\ref{tab:audio_results} summarizes the results. Interestingly, S2T KD combined with T2T KD enhances the model's ability for acoustic analysis and reasoning over human speech. Second, incorporating additional AQA data (6181 samples from Clotho \cite{Drossos2019}) leads to simultaneous improvements in the Sound and Speech categories. This suggests that future work could explore combining semantic dialogue knowledge distillation with acoustic audio analysis distillation.

\section{Conclusion}
We study catastrophic forgetting and modality inequivalence in speech LLMs, showing that adding speech capabilities degrades text knowledge, especially with spoken queries. To address this, we propose a cross-modal distillation framework combining text and speech channels. Experiments on dialogue and audio QA tasks show our method preserves textual knowledge, improves cross-modal alignment, and enhances speech-based reasoning, enabling more robust speech LLMs.

\newpage

\bibliographystyle{IEEEbib}
\bibliography{strings,refs}

\end{document}